# Sufficient Dimensionality Reduction with Irrelevance Statistics

Amir Globerson*　　　Gal Chechik *　　　Naftali Tishby

School of Computer Science and Engineering and
Interdisciplinary Center for Neural Computation
The Hebrew University, Jerusalem 91904, Israel

## Abstract

The problem of unsupervised dimensionality reduction of stochastic variables while preserving their most relevant characteristics is fundamental for the analysis of complex data. Unfortunately, this problem is ill defined since natural datasets inherently contain alternative underlying structures. In this paper we address this problem by extending the recently introduced "Sufficient Dimensionality Reduction" feature extraction method [7], to use "side information" about irrelevant structures in the data. The use of such irrelevance information was recently successfully demonstrated in the context of clustering via the *Information Bottleneck* method [1]. Here we use this side-information framework to identify continuous features whose measurements are maximally informative for the main data set, but carry as little information as possible on the irrelevance data set. In statistical terms this can be understood as extracting statistics which are maximally sufficient for the main dataset, while simultaneously maximally ancillary for the irrelevance dataset. We formulate this problem as a tradeoff optimization problem and describe its analytic and algorithmic solutions. Our method is demonstrated on a synthetic example and on a real world application of face images, showing its superiority over other methods such as Oriented Principal Component Analysis.

## 1 INTRODUCTION

Extracting regular structures or compact features from empirical data is a fundamental problem in machine learning. Such structures are essential for any comprehensible model of the data or for making useful predictions. Unfortunately, natural datasets inherently contain alternative, often conflicting, underlying structures. Thus for instance, spoken utterances can be labeled by their contents or by the speaker's identity; face images can be categorized by either a person's identity or expression; protein molecules can be classified by their physical structure or biological function. All are valid alternatives for analyzing the data, but the question of the "correctness" or "relevance" depends on the task. The "noise" in one analysis is the "signal" for another.

A novel approach to this old problem was recently proposed in the "Information Bottleneck with Side Information" (IBSI) method [1], by clustering (compressing) one variable in a way that keeps the (mutual) information about another variable (the "Information Bottleneck"), while utilizing additional *irrelevance* data as "side information". Such irrelevance information is very often available in terms of joint statistics of our variables in another context, but the irrelevant attributes are usually not explicit. A typical example is the analysis of gene expression data for some pathology, where the irrelevance information can be given in terms of the expression of control, healthy, tissues. In this case it is essentially impossible to isolate the irrelevant variables, though they are implicitly expressed in the expression patterns and statistics. The goal of the new unsupervised learning algorithms is to identify structures which are characteristic to the relevant dataset, but do not describe well the irrelevance data. The nature of the structures may vary: In [1, 15] they are represented by clusters, namely discrete quantization of the compressed variable. In other applications, continuous features are more appropriate than quantization (e.g. in [12, 13]).

We have recently introduced an information theoretic notion of continuous structure extraction in contingency tables, called "Sufficient Dimensionality Reduc-tion" (SDR) [7]. This method aims at extracting continuous features that are "approximate sufficient statistics" for a variable $X$ whose expected value provides maximal information about another, relevance, variable $Y$. The SDR algorithm finds such informative features by searching for the best exponential form approximation to the given joint distribution. It is a principled nonlinear dimension reduction method which favorably competes with other dimensionality reduction algorithms.

---

* Both authors contributed equally.

In this paper we generalize SDR to take advantage of irrelevance information. Our method, *SDR with Irrelevance Statistics* (SDR-IS), finds features which are maximally informative about one, relevant, variable $Y^+$, while being minimally informative about another one $Y^-$ provided as irrelevance information. Once the question is properly posed using information theoretic measures it is merely the power of the principled formulation of SDR and IBSI problems that yields the solution to the new problem.

Features (i.e. statistics of empirical data) that carry no information about a parameter are known as ancillary statistics [6]. These are mainly used for estimating precision of the standard estimators. SDR-IS extracts features that are approximately sufficient for the relevant variable $Y^+$, and in the same time approximately ancillary for $Y^-$. The quantitative nature of the approximation is determined by a trade-off between the information that the extracted features carry about $Y^+$ and the information they maintain about $Y^-$.

The next two sections formalize the problem of continuous feature extraction with irrelevance information using the notion of "information in a measurement". We then derive its formal (section 4) and algorithmic (section 5) solutions, relate them to likelihood ratio maximization (section 6), and demonstrate their operation on synthetic and real world problems (section 7).

## 2   PROBLEM FORMULATION

To formalize the above ideas consider a scenario where two empirical joint distributions are given for three categorical random variables $X, Y^+$ and $Y^-$. The first is the main data, $p^+ \equiv P(X, Y^+)$, which describes the joint distribution of $Y^+$ and $X$. The second is the irrelevance data, $p^- \equiv P(X, Y^-)$, which is assumed to contain irrelevant structures in the main data. Our goal is to identify features of $X$ that characterize its probabilistic relation to $Y^+$ but not its relation to $Y^-$. Note that $Y^+$ and $Y^-$ need not come from the same space, or have the same size and dimension. Potentially, one may be continuous and the other discrete, although we do not treat the continuous case here.

We seek a $d$ dimensional continuous feature of $X$ which we denote $\vec{\phi}(x) : X \to \Re^d$, such that only its expected values $\langle \vec{\phi}(x) \rangle_{p(x|y^+)}$ characterize the stochastic dependence between $X$ and $Y^+$, while the corresponding values for $Y^-$, namely $\langle \vec{\phi}(x) \rangle_{p(x|y^-)}$, do not characterize the dependence of $Y^-$ on $X$. For example, the mean number of words in some semantic set may reveal a document's writing style, but tell us nothing of its content. Here, $Y^+$ would be a set of documents of different writing styles, and $Y^-$ a set of documents with the same style but varying contents. $X$ will represent the set of words.

The idea of using expected values of features to describe a distribution stands in the basis of the Maximum-Entropy (MaxEnt) approach [8, 4]. On one hand, these descriptions provide a natural way to efficiently estimate and represent distributions using parametric representations. Furthermore, the extracted parameters often provide compact description of the data in terms of interpretable features. While in standard MaxEnt the features are predetermined (or greedily optimized over a given set as in [4]), we have presented in [7] a method for finding features which are optimal for a given set of distributions over $X$, thus solving an *Inverse Maximum Entropy* problem. The continuous features $\vec{\phi}(x)$ can be any $d$ dimensional function of a discrete variable $X$. To evaluate the "goodness" of $\vec{\phi}(x)$, we use the notion of *measurement information* $I_M[\vec{\phi}(x), p]$, defined in [7] and reviewed in the next section. Given this measure for the quality of $\vec{\phi}(x)$, the goal of relevant feature extraction is to identify features that are maximally informative about $Y^+$ while minimally informative about $Y^-$. This dual optimization task can be approached by minimizing the weighted difference

$$\mathcal{L}[\vec{\phi}(x)] = I_M[\vec{\phi}(x), p^+] - \lambda I_M[\vec{\phi}(x), p^-] \quad (1)$$

over $\vec{\phi}(x)$, where $\lambda$ is a positive tradeoff parameter reflecting the weight to be assigned to the irrelevance data.

## 3   INFORMATION IN MEASUREMENTS

Given a joint distribution $p(X, Y)$, we define the *measurement* of $\vec{\phi}(x)$ as the set of expected values $\langle \vec{\phi}(x) \rangle_{p(x|y)}$ and its marginals $p(x), p(y)$. In order to find optimal features, one needs to quantify how well such measurements capture the information between $X$ and $Y$. Note that this information is *not* quantified by the Shannon mutual information $I(\vec{\phi}(X); Y)$ which is equal to $I(X; Y)$ for discrete variables. It is also not captured by $I(\sum_{i=1}^n \vec{\phi}(X_i); Y)$, where $X_1, \ldots, X_n$ is an *i.i.d.* sample of $X$ given a value of $Y$, since as $n$ goes to infinity this empirical mean converges to its expected value, and the information measure degenerates to the entropy of $Y$.

The approach taken in [7] builds on the *MaxEnt* philosophy, by searching for the distribution which has the same measurement values as $p(X,Y)$ but contains minimal mutual information between $X$ and $Y$. This information effectively extracts the dependence between $X$ and $Y$ which can be attributed to the knowledge of $\vec{\phi}(x)$ and its expectations.

Formally, denote the set of these distributions by

$$\mathcal{P}(\vec{\phi}(x),p) \equiv \left\{ \hat{p}(X,Y) : \begin{array}{c} \langle\vec{\phi}(x)\rangle_{\hat{p}(x|y)} = \langle\vec{\phi}(x)\rangle_{p(x|y)} \\ \hat{p}(x) = p(x) \\ \hat{p}(y) = p(y) \end{array} \right\}. \quad (2)$$

We define the information in the measurement of $\vec{\phi}(x)$ as

$$I_M[\vec{\phi}(x),p] \equiv \min_{\hat{p}(X,Y)\in\mathcal{P}(\vec{\phi}(x),p)} I[\hat{p}] \quad (3)$$

where $I[p]$ is the Shannon mutual information of the two variables $X$ and $Y$ with joint distribution $p(X,Y)$ [2] [1]

$$I[p(X,Y)] \equiv \sum_{x,y} p(x,y) \log \left( \frac{p(x,y)}{p(x)p(y)} \right) . \quad (4)$$

The optimization problem in Equation 1 thus becomes:

$$\begin{aligned} \vec{\phi}^*(x) &= \arg\max \mathcal{L}[\vec{\phi}(x)] \quad (5) \\ &= \arg\max_{\vec{\phi}(x)} \min_{\hat{p}^+\in\mathcal{P}(\vec{\phi},p^+)} I[\hat{p}^+] - \lambda \min_{\hat{p}^-\in\mathcal{P}(\vec{\phi},p^-)} I[\hat{p}^-] \end{aligned}$$

## 4 SOLUTION CHARACTERIZATION

In order to characterize the solution of the variational problem in Equation 5, we now calculate its gradient and observe its vanishing points. We start by characterizing the form of the distribution $\hat{p}_\phi(X,Y)$ that achieves the minimum of $I_M[\vec{\phi}(x),p]$ (Equation 3). Since $I[\hat{p}(X,Y)] = H[\hat{p}(X)] + H[\hat{p}(Y)] - H[\hat{p}(X,Y)]$, and the marginals $\hat{p}(X)$, $\hat{p}(Y)$ are kept constant by the definition of $\mathcal{P}(\vec{\phi}(x),p)$, we have $I[\hat{p}(X,Y)] = const - H[\hat{p}(X,Y)]$. This turns Equation 3 into a problem of entropy maximization under linear constraints

$$\hat{p}_\phi(X,Y) = \max_{\hat{p}(X,Y)\in\mathcal{P}(\vec{\phi}(x)),p)} H[\hat{p}(X,Y)] , \quad (6)$$

---

[1] We use here the notation $I[p]$ instead of the more common notation $I(X;Y)$ to emphasize that $I$ is a functional of the distribution $p$. We also use $H[p]$ to denote the entropy of $p$.

whose solutions are known to be of exponential form [4]

$$\hat{p}_\phi(x,y) = \frac{1}{Z} \exp\left(\vec{\phi}(x)\cdot\vec{\psi}_\phi(y) + A_\phi(x) + B_\phi(y)\right). \quad (7)$$

The $\vec{\psi}_\phi(y), A_\phi(x)$ and $B_\phi(y)$ are complex functions of $\vec{\phi}(x)$ that play the role of Lagrange multipliers in the maximum entropy problem derived from Equation 6.

While $H[\hat{p}_\phi(X,Y)]$ is a complex function of $\vec{\phi}(x)$, its gradient can be derived analytically using the fact that $\hat{p}_\phi$ has the exponential form of Equation 7. Appendix A shows that this gradient is

$$\frac{\partial H[\hat{p}_\phi(X,Y)]}{\partial \vec{\phi}(x)} = p(x)\left(\langle\vec{\psi}_\phi\rangle_{\hat{p}_\phi(y|x)} - \langle\vec{\psi}_\phi\rangle_{p(y|x)}\right) \quad (8)$$

It is now straightforward to calculate the gradient of the functional in Equation 5. Denote by $\hat{p}_\phi^+$ and $\hat{p}_\phi^-$ the information minimizing distributions obtained in $I_M[\phi,p^+]$ and $I_M[\phi,p^-]$, and by $\vec{\psi}_\phi^+$ and $\vec{\psi}_\phi^-$ their corresponding Lagrange multipliers. The gradient is then

$$\begin{aligned} \frac{\partial \mathcal{L}}{\partial \vec{\phi}(x)} &= p^+(x)\left(\langle\vec{\psi}_\phi^+\rangle_{p^+(y^+|x)} - \langle\vec{\psi}_\phi^+\rangle_{\hat{p}_\phi^+(y^+|x)}\right) \quad (9) \\ &\quad -\lambda p^-(x)\left(\langle\vec{\psi}_\phi^-\rangle_{p^-(y^-|x)} - \langle\vec{\psi}_\phi^-\rangle_{\hat{p}_\phi^-(y^-|x)}\right) \end{aligned}$$

Setting it to zero we obtain the characterization of the extremum point

$$p^+(x)\Delta\langle\vec{\psi}_\phi^+\rangle = \lambda p^-(x)\Delta\langle\vec{\psi}_\phi^-\rangle \quad (10)$$

where $\Delta\langle\psi\rangle$ is the difference in the expectation of $\psi$ taken according to the model and the true distribution.

To obtain some intuition into the last equation consider the following two observations. First, note that maximizing the information $I_M[\phi,p^+]$ requires to minimize the absolute difference between the expectancies of $\vec{\psi}_\phi^+$, as can be seen when taking $\lambda = 0$. Second, it can be shown than when minimizing $I_M[\phi,p^-]$ alone, some elements of $\vec{\phi}(x)$ must diverge. In these infimum points $\Delta\langle\vec{\psi}_\phi^-\rangle$ does not generally vanish. Taken together, these facts imply that for the $\lambda > 0$ case, the difference $\Delta\langle\vec{\psi}_\phi^+\rangle$ should generally be different from zero. This implies, as expected, that the resulting $\vec{\phi}(x)$ conveys less information than the $\lambda = 0$ solution. The optimal $\vec{\phi}(x)$ is thus bound to provide an inaccurate model for those aspects of $p^+$ that also improve the model of $p^-$.

An additional interesting interpretation of $\vec{\psi}_\phi^+$, $\vec{\psi}_\phi^-$ is that they reflect the relative importance of $\vec{\phi}(x)$ in $\hat{p}_\phi^+$, $\hat{p}_\phi^-$ for a given $y$. This view is prevalent in the boosting literature, where such coefficients function as the weights of the weak learners (see e.g. [9]). However, SDR-IS also optimizes the weak learners, searching for a small but optimal set of learners.

## 5 ALGORITHMIC CONSIDERATIONS

Unlike the case of $\lambda = 0$ for which an iterative algorithm was described [7], the $\lambda > 0$ case poses a special difficulty in developing such an algorithm. One could supposedly proceed by calculating $\vec{\psi}_\phi^+, \vec{\psi}_\phi^-$ assuming a constant value of $\vec{\phi}(x)$ and then calculate the resulting $\vec{\phi}(x)$ assuming $\vec{\psi}^+$ and $\vec{\psi}^-$ are constant. However, as was shown in [7] updating $\vec{\psi}_\phi^-$ will increase $I_M[\vec{\phi}(x), p^-]$ thereby decreasing the target function. Thus, such a procedure is not guaranteed to improve the target function. Possibly, an algorithm guaranteed to converge for a limited range of $\lambda$ values can be devised, as done for IBSI [1], but this remains to be studied.

Fortunately, the analytic characterization of the gradient derived above allows one to use a gradient ascent algorithm for finding the optimal features $\vec{\phi}(x)$, for any given value of $\lambda$. This requires to calculate a Maximum Entropy distribution on each of its iterations, namely, to calculate numerically the set of Lagrange multipliers $\vec{\psi}_\phi(y)$, $A_\phi(x)$ and $B_\phi(y)$ which appear in the gradient expression in Equation 8. This convex problem has a single maximum, and well studied algorithms exist for finding Maximum Entropy distributions under linear constraints [2]. These include GIS [3], IIS [4], or gradient based algorithms (see [10] for a review of different algorithms and their relative efficiency). In all the results described below we used the GIS algorithm.

## 6 RELATION TO OTHER METHODS

### 6.1 Likelihood Ratio Maximization

Further intuition into the functional of Equation 5 can be obtained, using the result of [7] yielding that it equals up to a constant to

$$\mathcal{L}[\vec{\phi}(x)] = -D_{KL}[p^+||\hat{p}_\phi^+] + \lambda D_{KL}[p^-||\hat{p}_\phi^-] \quad (11)$$

where $D_{KL}[p||q] \equiv \sum p_i \log(p_i/q_i)$ is the Kullback-Leibler divergence. When $p^+$ and $p^-$ share the same marginal distribution $p(x)$, a joint distribution $p(X, Y^+, Y^-)$ can be defined that coincides with the pairs-joint distributions $p^+(X, Y^+)$ and $p(X, Y^-)$,

$$p(x, y^+, y^-) \equiv p^+(y^+|x)p^-(y^-|x)p(x) \quad (12)$$

The above distribution has the quality that $Y^-$ and $Y^+$ are conditionally independent given $X$. In many

---

[2]Note that all the constraints in $\mathcal{P}(\vec{\phi}(x), p)$ are indeed linear.

settings, this is indeed a reasonable assumption. In this case

$$\begin{aligned}\mathcal{L} &= -\sum_{x,y^+,y^-} p(x,y^+,y^-) \log\left(\frac{p^+(x,y^+)}{\hat{p}_\phi^+(x,y^+)}\right) \quad (13) \\ &+ \lambda \sum_{x,y^+,y^-} p(x,y^+,y^-) \log\left(\frac{p^-(x,y^-)}{\hat{p}_\phi^-(x,y^-)}\right) \\ &= -\left\langle \log\left(\frac{p^+(x,y^+)}{p^-(x,y^-)^\lambda} \frac{\hat{p}_\phi^-(x,y^-)^\lambda}{\hat{p}_\phi^+(x,y^+)}\right) \right\rangle_{p(x,y^+,y^-)} \\ &= \left\langle \log\left(\frac{\hat{p}_\phi^+(x,y^+)}{\hat{p}_\phi^-(x,y^-)^\lambda}\right) \right\rangle_{p(x,y^+,y^-)} + const\end{aligned}$$

This suggests that in the special case of $\lambda = 1$, SDR-IS operates to maximize the expected log likelihood ratio, between the maximum entropy models $\hat{p}_\phi^+$ and $\hat{p}_\phi^-$. In the general case of $\lambda > 0$ a weighted log likelihood ratio is obtained. For vanishing $\lambda$, the irrelevance information is completely ignored and the problem reduces to unconstrained likelihood maximization of the maximum entropy model $\hat{p}_\phi^+$.

### 6.2 Weighted vs. Constrained Optimization

The trade-off optimization problem of Equation 1, is related to the following constrained optimization problem

$$\vec{\phi}^*(x) = \arg\max_{\vec{\phi}(x): I_M[\vec{\phi}(x), p^-] < D} I_M[\vec{\phi}(x), p^+] \quad . \quad (14)$$

Although the Lagrangian for this problem is identical to the SDR-IS target functional of Equation (1), these two problems are not necessarily equivalent, since a constrained optimization problem like (14) may in principle be solved by the minimum point of Equation 1. However, under certain convexity conditions such problems can be shown to be equivalent. While we do not present a similar proof here, we found numerically in all the data described below, that the maximum points of 14 were always achieved at the maximum of Equation (1) rather than at its minima.

### 6.3 Related Methods

Several methods previously appeared in the literature, which make use of auxiliary data or additional sources of information to enhance learning features of a main data set. The method of Oriented-PCA [5] uses a main data set with covariance $S^+$ and an irrelevance data set with covariance $S^-$ to find features $w$ that maximize the Signal to Noise Ratio $\frac{w^T S^+ w}{w^T S^- w}$. Constrained-PCA [5] finds principal components of the main data which are orthogonal to the irrelevance data. While these methods implicitly assume Gaussian distributions in input space, a kernelized version of OPCA was described in [12].

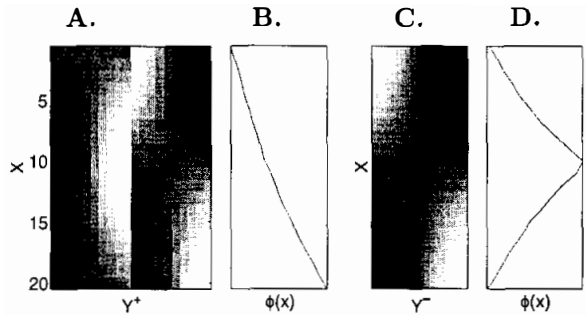

Figure 1: Demonstration of SDR-IS operation. **A.** A joint distribution $P(X, Y^+)$ that contains two distinct and conflicting structures (see text) **B.** Extracting a one-dimensional feature with $\lambda = 0$ identifies the top-to-bottom gradient. **C.** A joint distribution $P(X, Y^-)$ that contains a single structure similar to the right structure of $P(X, Y^+)$. **D.** Extracting a one-dimensional feature with $\lambda = 1$ successfully ignores the top-to-bottom gradient and extracts the weaker structure of $P(X, Y^+)$.

Another line of work uses auxiliary information in the form of equivalence constraints. The auxiliary data here is a set of relations that enforce similarity between the elements of the main data. These relations are used to improve dimensionality reduction [13], or to improve the distance metrics used for clustering [15].

Separating several conflicting structures in the data has also been addressed in [14] where a bilinear model was used to separate style from content. This model does not use auxiliary information, but rather assumes that the two structures can be represented by a linear model.

SDR-IS differs from the above methods in that it is a non-linear method for extracting continuous features, which are **least informative** about the irrelevance data. The relative importance of the irrelevance data is determined throught the tradeoff parameter $\lambda$.

## 7 APPLICATIONS

We first illustrate the operation of SDR-IS on a synthetic example that demonstrates its main properties. Then, we describe its application to the problem of feature extraction for face recognition.

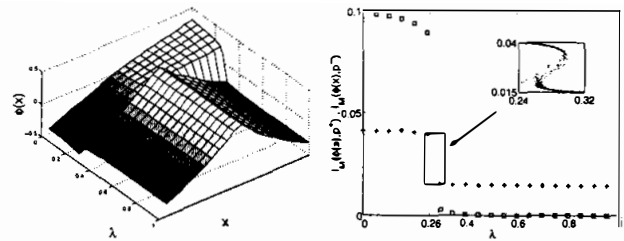

Figure 2: Operation of SDR-IS on the synthetic example of Figure 1 for various values of $\lambda$. **A.** The optimal $\vec{\phi}(x)$ extracted with SDR-IS. **B.** The information conveyed about $Y^+$ (crosses) and $Y^-$ (squares), by the optimal $\vec{\phi}(x)$'s of the left panel. A phase transition around 0.26 is observed both in the information values and the $\vec{\phi}(x)$'s. The inset shows the spinodal metastable points of $I_M[\vec{\phi}(x), p^+]$ around the phase transition point (black box).

### 7.1 A Synthetic Example

To demonstrate the ability of our approach to uncover weak but interesting hidden structures in data, we designed a co-occurrence matrix that contains two competing sub-structures (see figure 1A). The right half of the matrix contains a top-to-bottom gradient, while its left half contains large variance at the middle values of $X$. The right structure was hand-crafted to be stronger in magnitude than the left one.

When SDR-IS is applied with no irrelevance information ($\lambda = 0$) and $d = 1$, it extracts the top-to-bottom gradient (Figure 1B). This $\phi(x)$ follows from the strong structure on the right part of 1A.

We now created a second co-occurrence matrix $P(X, Y^-)$ that contains a top-to-bottom structure similar to that of $P(X, Y^+)$ (Figure 1C). Applying SDR-IS with $\lambda = 1$ on both matrices now successfully ignores the strong top-to-bottom structure in $P(X, Y^+)$ and retrieves the weaker structure that emphasizes the mid values of $X$ (Figure 1D). Importantly, this is done in an unsupervised manner, without explicitly pointing to the strong but irrelevant structure.

Further understanding of the operation of SDR-IS is gained by tracing its output as a function of the tradeoff parameter $\lambda$. Figure 2A plots the optimal features $\vec{\phi}(x)$ extracted for various $\lambda$ values, revealing a phase transition around a critical value $\lambda = 0.26$. The reason for this behavior is that at the critical $\lambda$, the top-to-bottom feature $\vec{\phi}(x)$ bears larger loss (due to the information $I_M[\vec{\phi}(x), p^-]$ conveyed about $Y^-$) than gain. Figure 2B traces the values of $I_M[\vec{\phi}(x), p^-]$ and $I_M[\vec{\phi}(x), p^+]$, again revealing a pronounced phase transition, and an S shaped (spinodal) curve of $I_M[\vec{\phi}(x), p^+]$, indicating the co-existence of three local maxima and the metastable region (inset of Figure 2B). Such spinodal curves are typical to phase transition phenomena obeserved in numerous physical systems. Plotting the SDR-IS functional of Equation 1 as a function of $\lambda$ (not shown) also reveals a discontinuity in its first derivative, indicating a first order phase transition. These discontinuities reflect the removal of "irrelevant" features from $\vec{\phi}(x)$, and can thus be used to select interesting values of $\lambda$.

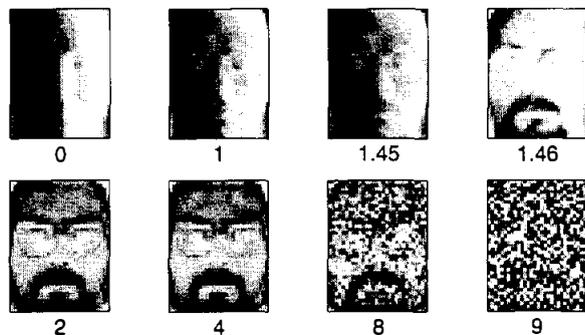

Figure 3: Extracting a single feature using SDR-IS, for various $\lambda$ values. An apparent phase transition is observed around $\lambda = 1.45$. $p^+$ was created by taking images of all men in the AR database with neutral face expressions and light either from the right or the left (a total of 100 images). $p^-$ was similarly created with 100 female images. Positive $\lambda$ values reveal features that differentiate between men but not between women.

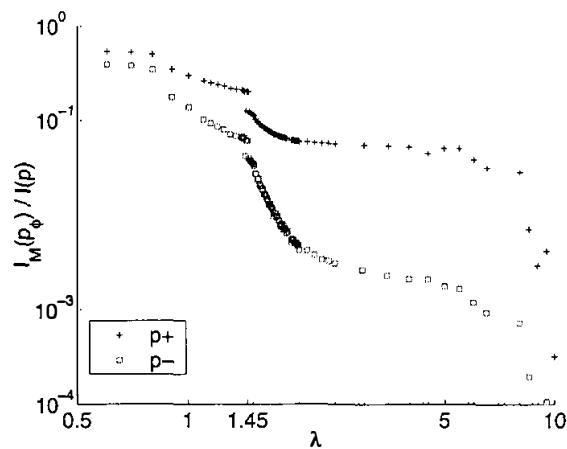

Figure 4: Normalized information about the main data $I_M[\vec{\phi}(x), p^+]$ and the irrelevance data $I_M[\vec{\phi}(x), p^-]$, as a function of $\lambda$, for the data of Figure 3. Note the phase transition in both information levels for $\lambda = 1.45$.

The irrelevant structures in the above example were hand crafted to be strongly and cleanly manifested in $P(X; Y^-)$. The next section studies the application of SDR-IS to real data, in which structures are much more covert.

### 7.2 Face Images

Face recognition poses a challenge to relevant features extraction since these must be invariant to various interfering structures, such as face expression and light conditions. Such nuisance structures are often more pronounced in the data than the subtle features required to recognize a person.

We tested SDR-IS on this task using the AR database [11], a collection of faces with various face expression light conditions and occlusions. Each image was translated into a joint probability matrix, by considering the normalized grey levels of the pixel $x$ in the image $y$ as the probability $p(x|y)$, and setting $p(y)$ uniform. This normalization scheme views every image $y$ as a distribution $p(x|y)$ which stands for the probability of observing a photon at a given pixel $x$. To demonstrate the operation of SDR-IS on this data we first trained it to extract a single feature, for various $\lambda$ values. The experiment details and resulting $\vec{\phi}(x)$ are given in Figure 3. When $\lambda$ is low (small weight for irrelevance information) the main structure captured is the direction of light source (right vs. left). As $\lambda$ increases the optimal $\vec{\phi}(x)$ first changes only slightly, but then a phase transition occurs around $\lambda = 1.45$, and a second structure emerges. This phase transition can be well observed when tracing the values of $I_M[\vec{\phi}(x), p^+]$ and $I_M[\vec{\phi}(x), p^-]$ as a function of $\lambda$ (Figure 4), and results from the same reasons discussed in the synthetic example described above. This result suggests that such information curves can be used to identify "interesting" values of $\lambda$ and their corresponding features even for high dimensional and complex data. As $\lambda$ further increases, the algorithm focuses on minimizing information about the irrelevance information, disregarding information about the main data. This results in the noisy features seen in figure 3 for high $\lambda$ values.

To quantify the performance of SDR-IS in a comparative manner, we used it in a difficult task of unsupervised feature extraction for face recognition, and compared its performance with three methods: PCA - the most widely used dimensionality reduction method; Constrained PCA (CPCA); and oriented PCA (OPCA) - two methods that utilize the same irrelevance data as SDR-IS [5]. We created $p(X, Y^+)$ with images of five different men, under all the different conditions of face expression and light conditions (a total of 26 images per person). As irrelevance data we used all 26 images of another randomly chosen man. The task of clustering these images into the five correct sets is hard since the nuisance structures are far more dominant than the relevant structure of inter subject variability, in face of light and face expression invariances.

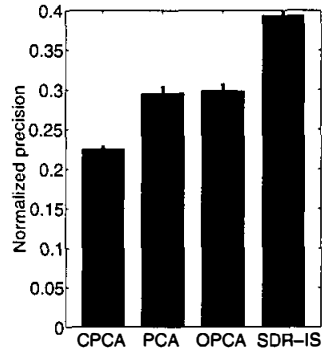

Figure 4. Performance of SDR-IS compared with other methods. Performance is normalized between 0 (obtained with random neighboring) and 1 (all nearest neighbors are of the same class). The average over ten cross validation sets is shown. SDR-IS achieves 30 percent improvement over the second best method (OPCA).

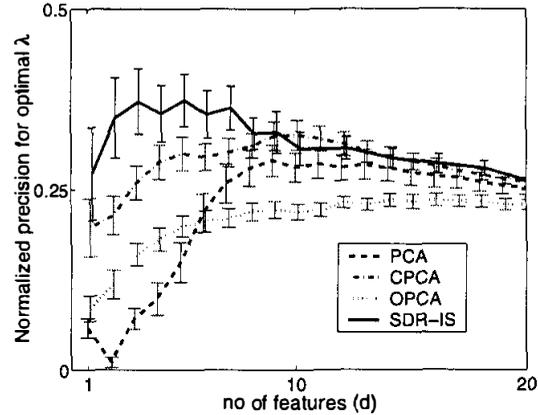

Figure 5: Performance of SDR-IS compared with other methods as a function of dimensionality $d$ for the AR data. The mean performance over 10 testing sets is reported, and bars denote standard error of the mean over these sets. In SDR-IS, a value of $\lambda$ was chosen for each $d$, to maximize performance over the training set

All methods were used to reduce the dimensionality of the images. PCA representations were obtained by projecting on the principal components. The SDR-IS representation was obtained by replacing each image $y$ with its expected SDR-IS feature values $\langle \vec{\phi}(x) \rangle_{p(x|y)}$. This follows our motivation of using expected values alone to represent $y$.

To quantify the effectiveness of the reduced representations in preserving person identity, we calculated the number of same-class (same-person) neighbors out of the $k$ nearest neighbors of each image[3]. This was averaged over all images and $k$'s and normalized, yielding the precision index [4].

Optimal parameters (dimensionality and $\lambda$) for all methods, were chosen to maximize the precision index for a training set. Reported results were obtained on a separate testing set. This entire procedure was repeated 10 times on randomly chosen subsets of the database. Figure 4 compares the effectiveness of SDR-IS with the one obtained with PCA based methods. SDR-IS was found to achieve more than 30 percent improvement over the second best method.

We further compared the performance of the four methods for each predefined dimensionality $d$. Figure 5 shows that SDR-IS dominates the other methods over all $d$ values. This is more pronounced for low values of $d$, which agrees with the intuition that the irrelevance data allows SDR-IS to focus on the more relevant features.

## 8 Discussion

The method introduced in this paper addresses the fundamental problem of extracting relevant structure in an unsupervised manner, a problem for which only few principled approaches were suggested. We focus on continuous features of categorical variables and use the information theoretic notion of *information in the expectation value of a measurement*, to derive algorithms that extract the most informative features, by utilizing information about irrelevant properties. Such an information theoretic approach makes no assumptions about the origin of the empirical data and is thus different from the more common generative modeling methodology.

Our formalism can be extended to the case of multiple relevance and irrelevance variables $(Y^+_1, ..., Y^+_{n+})$ and $(Y^-_1, ..., Y^-_{n-})$, with joint distributions $p^+_i \equiv p(X, Y^+_i)$ and $p^-_i \equiv p(X, Y^-_i)$. Following a similar weighted optimization problem we write the Lagrange form of the functional $\mathcal{L} = \sum_{i=1}^{n^+} \lambda_i I_M[\vec{\phi}(x), p^+_i] - \sum_{i=1}^{n^-} \lambda_i I_M[\vec{\phi}(x), p^-_i]$, which can be maximized as in the two variables case.

An interesting property that is revealed when applying SDR-IS both to synthetic and real life data, is the emergence of phase-transitions. These are discontinuous changes in the information values, that occur at specific values of $\lambda$. They are paralleled by abrupt changes in the features $\vec{\phi}(x)$ and thus provide a natural way to focus on important values of $\lambda$ that characterize the inherent features of the data. As demonstrated in Figure (4), we were able to follow the metastable re-gion of the phase transition, which appears to behave like a first-order transition in thermodynamics.

---

[3] As a metric for measuring distances between images, we tested both the L2 norm and the Mahalanobis distance in the reduced representation. We report the Mahalanobis results only, since L2 results were considerably worse for PCA.

[4] We also evaluated the methods by clustering the low dimensional vectors into five groups and comparing the resulting clusters with the true ones. This resulted in qualitatively similar result, albeit noisier. We prefer the method presented here since it does not depend on a noisy second phase of clustering.

An interesting algorithmic problem, not fully answered at this point, is to design an iterative-projection algorithm, similar to the SDR, for solving the implicit equations for the optima. This can improve time complexity and convergence of the algorithm, making it even more practical.

## Acknowledgements

We thank R. Gilad-Bachrach and A. Navot for helpful discussions. A.G. and G.C. are supported by the Israeli ministry of Science, the Eshkol fellowship. This work is partially supported by a grant from the Israeli Academy of Science (ISF).

## A  DERIVING THE GRADIENT OF THE JOINT ENTROPY

To calculate the gradient of the entropy $H[\hat{p}_\phi(x,y)]$, we first prove three useful properties of the distribution $\hat{p}_\phi$. Since $\hat{p}_\phi$ is in $\mathcal{P}(\vec{\phi}(x),p)$, it satisfies the margin constraints: $\hat{p}_\phi(x) = \sum_{y'} \hat{p}_\phi(x,y') = p(x)$, $\hat{p}_\phi(y) = \sum_{x'} \hat{p}_\phi(x',y) = p(y)$, as well as the expectation constraints $\sum_{x'} \vec{\phi}(x')(\hat{p}_\phi(x',y) - p(x',y)) = 0$. Deriving the three constraints equations w.r.t. $\vec{\phi}(x)$ yields

$$\sum_{y'} \frac{\partial \hat{p}_\phi(x,y')}{\partial \vec{\phi}(x)} = 0; \quad \sum_{x'} \frac{\partial \hat{p}_\phi(x',y)}{\partial \vec{\phi}(x)} = 0 \quad (15)$$

for the marginal constrains, and

$$\hat{p}_\phi(x,y) - p(x,y) + \sum_{x'} \vec{\phi}(x') \frac{\partial \hat{p}_\phi(x',y)}{\partial \vec{\phi}(x)} = 0 \quad (16)$$

for the expectation constraints.

The derivative of the entropy can now be written as

$$\begin{aligned}\frac{\partial H[\hat{p}_\phi]}{\partial \vec{\phi}(x)} &= -\sum_{x',y'} \frac{\partial \hat{p}_\phi(x',y')}{\partial \vec{\phi}(x)} & (17)\\&\quad -\sum_{x',y'} \frac{\partial \hat{p}_\phi(x',y')}{\partial \vec{\phi}(x)} \log \hat{p}_\phi(x',y') \\&= -\sum_{x',y'} \frac{\partial \hat{p}_\phi(x',y')}{\partial \vec{\phi}(x)} \log \hat{p}_\phi(x',y')\end{aligned}$$

where the last equality stems from the vanishing derivative of the marginal constraints in Equation 15. Plugging in the exponential form of $\hat{p}_\phi$ from Equation 7, and using Equation 15 again, we have

$$\frac{\partial H[\hat{p}_\phi]}{\partial \vec{\phi}(x)} = -\sum_{x',y'} \frac{\partial \hat{p}_\phi(x',y')}{\partial \vec{\phi}(x)} \vec{\phi}(x') \cdot \vec{\psi}_\phi(y') \quad (18)$$

Now using Equation 16 for the derivative of the expectation constraints, we finally obtain

$$\frac{\partial H[\hat{p}_\phi]}{\partial \vec{\phi}(x)} = p(x) \left( \langle \vec{\psi}_\phi \rangle_{\hat{p}_\phi(y|x)} - \langle \vec{\psi}_\phi \rangle_{p(y|x)} \right) . \quad (19)$$